\documentclass[twocolumn]{article}

\usepackage[width=17cm,height=22cm]{geometry}
\usepackage[utf8]{inputenc}
\usepackage{fancyvrb}
\usepackage{authblk}
\usepackage{hyperref}
\usepackage{graphicx}
\usepackage{amssymb,amsmath,amsfonts}  
\usepackage{csquotes}
\usepackage{listings}
\usepackage{hyperref}

\title{DistilCamemBERT : une distillation du modèle français CamemBERT}
\author[1]{Cyrile Delestre\thanks{cyrile.delestre@arkea.com}}
\author[1]{Abibatou Amar\thanks{amar.abibatou@gmail.com}}
\affil[1]{Crédit Mutuel Arkéa, 32 Rue Mirabeau, 29480 Le Relecq-Kerhuon, France}
\date{}

\begin{document}
\maketitle

\begin{abstract}
Les modélisations de \textit{Natural Language Processing} (NLP) modernes à base de structure \textit{Transformer} représentent l'état de l'art en terme de performances sur des tâches très diversifiées. Cependant ces modélisations sont complexes et représentent plusieurs centaines de millions de paramètres pour les plus modestes d'entre elles. Ce constat peut nuire à leur adoption au niveau industriel rendant la mise à l'échelle sur une infrastructure raisonnable et/ou conforme aux responsabilités sociétales et environnementales difficile. Dans ce but nous présentons dans cet article une modélisation permettant de réduire drastiquement le coût calculatoire d'un modèle français très connu (CamemBERT), tout en préservant de bonnes performances.
\end{abstract}

\medskip

\noindent\textbf{Mots-clef}: Distillation, CamemBERT, Transformers, NLP.

\section{Introduction}
\label{sec:intro}

C'est en juin 2017, que l'article fondateur de tout un pan des modélisations NLP est publié par Google, introduisant la structure Transformer~\cite{VaSh2017}, et en juin 2018 que la première modélisation réalisée par OpenAI reprenant cette structure est publiée sous le nom de \textit{Generative Pre-Traning} (GPT), utilisant une variante auto-régressive~\cite{RaNa2018}. Puis, c'est en octobre de cette même année que Google introduit la modélisation \textit{Bidirectional Encoder Representations from Transformers} (BERT)~\cite{DeCh2018}. Finalement, en juillet 2019, Meta (anciennement Facebook) introduit une variante sur la tokenization et la stratégie d’apprentissage offrant de meilleures performances sur les tâches de l'état de l'art, \textit{Robustly Optimized BERT pretraining Approach} (RoBERTa)~\cite{LiOtt2019}. Le français n'est pas en reste avec deux modélisations. En mai 2020, une nommée FlauBERT~\cite{JaVi2020}, est adaptée de BERT sur un corpus français et en juillet 2020, CamemBERT~\cite{LoMu2019} est quant à elle une adaptation de RoBERTa. Ces modélisations, bien que très performantes, sont très coûteuses en temps d'inférence. C'est dans l'optique de réduire cette contrainte que HuggingFace, en octobre 2019, introduit DistilBERT~\cite{SaDe2019}, une stratégie de distillation de la modélisation BERT afin de diviser le temps d'inférence par deux en réduisant la taille de la structure du réseau tout en limitant les pertes de performances sur les tâches de NLP. Cette stratégie d'apprentissage a donné toute une série de modélisations Distil* dans diverses langues, mais malheureusement pas en Français. C'est pour cette raison que nous introduisons aujourd'hui DistilCamemBERT, une distillation du modèle CamemBERT.

Dans une première partie nous aborderons les principes d'attention des réseaux Transformers afin de faciliter la description du principe de distillation sur ce type de modélisation dans une seconde partie. Puis, finalement, dans la troisième et dernière partie nous nous pencherons sur l'étude de performance de notre modèle DistilCamemBERT au travers de divers benchmarcks permettant de le confronter à d'autres types de modélisation dans des utilisations typiques. 

\section{Transformers}
\label{sec:transformers}

Dans cette section nous allons introduire la structure Transformer. Pour des raisons de simplicité nous omettrons volontairement les normalisations et les boucles résiduelles afin de faciliter l'explication et alléger les écritures. Nous définirons par la matrice $\mathbf{H}\in\mathbb{R}^{d_\mathbf{h}\times n}$ la représentation latente des tokens en sortie d'une couche Transformer où $d_\mathbf{h}$ représente la dimension latente des tokens et $n$ le nombre de tokens que compose la/le phrase/paragraphe/texte en entrée de la modélisation. Ainsi en sortie du $l$-ième Transformer nous avons $\mathbf{H}_l=[\mathbf{h}_{l,1}, \dots, \mathbf{h}_{l,n}]$.

La première représentation $\mathbf{H}_0$ en entrée de la première couche de Transformer est la concaténation des \textit{word embeddings} sommés avec l'encodeur de position qui permet d'ajouter \textit{une texture} au mot en fonction de sa position dans la phrase. Il est à noter que le \textit{positional encoder} est une fonction prédéfinie dans l'article original, alors que ce sont des coefficients appris par le réseau durant la phase d’entraînement pour leurs déclinaisons en BERT et RoBERTa.

Ainsi la $l$-ième couche Transformer peut être écrite comme suit :
\begin{equation}
\mathbf{H}_l = \mathrm{Transformer}_l(\mathbf{H}_{l-1}),\; l\in[\![1,L]\!]
\end{equation}

Les réseaux de neurones de type Transformers sont en réalité des structures relativement simples qui s'articulent autour d'une mémoire fonctionnelle. C'est-à-dire une mémoire qui permet de préserver la dérivation des fonctions composées rétro-propageant le gradient dans le réseau. C'est cette partie qui constitue la mécanique d'attention (\textit{Head Attention}).

Ainsi, à chaque couche, une représentation d'une requête (\textit{Query}), d'une clef (\textit{Key}) et d'une valeur (\textit{Value}) sont déterminées par les transformations linéaires suivantes :
\begin{eqnarray}
\mathbf{Q}_{j,l}(\mathbf{H})\!\! &\!\! = \!\!& \!\!\mathbf{W}_{j,l}^Q\mathbf{H}+\mathbf{1}_{1\times n}\otimes\mathbf{b}_l^Q\\
\mathbf{K}_{j,l}(\mathbf{H})\!\! &\!\! =\!\! &\!\!\mathbf{W}_{j,l}^K\mathbf{H}+\mathbf{1}_{1\times n}\otimes\mathbf{b}_l^K\\
\mathbf{V}_{j,l}(\mathbf{H})\!\! &\!\! =\!\! &\!\!\mathbf{W}_{j,l}^V\mathbf{H}+\mathbf{1}_{1\times n}\otimes\mathbf{b}_l^V
\end{eqnarray}
avec $\mathbf{W}_{j,l}^k\in \mathbb{R}^{\tfrac{d_\mathbf{h}}{J}\times d_\mathbf{h}}\;\forall k\in\{Q, K, V\}$, $J$ est le nombre de \textit{head attention} mis en parallèle à chaque couche d'attention (resp. $j$ est le $j$-ième), $\mathbf{1}_{1\times n}$ est un vecteur ligne de $n$ 1 et $\otimes$ est le produit de Kronecker.

Ainsi, plus une requête va concorder avec une clef par relation de colinéarité, plus la valeur émise associée à cette clef sera importante. Cette mécanique est modélisée de cette manière~:
\begin{equation}
\mathrm{Head}_{j,l}(\mathbf{H}) = \mathbf{V}_{j,l}\; \mathrm{softmax}\!\left(\frac{\mathbf{Q}_{j,l}^T\mathbf{K}_{j,l}}{\sqrt{d_\mathbf{h}/J}}\right)^T
\end{equation}
\noindent où l'opérateur $\mathrm{softmax}$ est calculé par rapport à la dimension des clefs (dernière dimension), permettant de répartir la probabilité des requêtes sur l'ensemble des clefs, et le rapport $\sqrt{d_\mathbf{h}/J}$ peut être vu comme un facteur de limite d’échelle permettant de maîtriser la dynamique du produit scalaire ($\tfrac{\mathbf{q}^T\mathbf{k}}{\sqrt{d_\mathbf{h}/J}}\underset{d_\mathbf{h}/J\to\infty}{\overset{\mathcal{L}}{\longrightarrow}}Z\sim\mathcal{N}(0,1)$) afin d'améliorer la rétro-propagation du gradient dans l'opérateur $\mathrm{softmax}$ (car des grandes valeurs entraînent de petits gradients).

A l'aide d'une telle mécanique il est facile de comprendre que deux éléments peuvent interagir entre eux sans être à proximité, \textit{a contrario} des méthodes de détection d'attention utilisant des réseaux de neurones récurrents, comme c'est le cas pour ELMo~\cite{MaMa2018}. De même l’interaction peut être indifféremment causale ou non vis-à-vis de la séquence de tokens, ce qui justifie le terme \textit{bidirectionnal} dans l’acronyme BERT.

Comme sous-entendu précédemment avec la valeur $J$, plusieurs mécaniques d'attention sont déployées en parallèle à chaque couche de Transformer, c'est ce qui est nommé le \textit{Multi-Head Attention}. Il s'agit de la simple concaténation des différentes attentions qui sont remélangées linéairement afin de revenir à une signification homogène de représentation latente par token~:
\begin{eqnarray}
\mathrm{MultiHead}_l(\mathbf{H})\!\! &\!\! =\!\! &\!\! \mathbf{W}_l^{out}\left[\mathrm{Head}_{1,l}^T, \dots, \mathrm{Head}_{J,l}^T\right]^T \nonumber\\
\!\! &\!\!\!\! &\!\! +\mathbf{1}_{1\times n}\otimes\mathbf{b}_l^{out}
\end{eqnarray}
\noindent avec $\mathbf{W}_l^{out}\in\mathbb{R}^{d_\mathbf{h}\times d_\mathbf{h}}$ .

Finalement la couche de Transformer se concrétise par un dernier mélange non-linéaire nommé réseau \textit{feed-forward} :
\begin{equation}
\mathrm{Transformer}_l(\mathbf{H}) = FF_l(\mathrm{MultiHead}_l(\mathbf{H}))
\end{equation}
avec
\begin{eqnarray}
FF_l(\mathbf{H})\!\! &\!\! =\!\! &\!\!\mathbf{W}_l^{forward}\mathrm{ReLu}(\mathbf{W}_l^{feed}\mathbf{H}+\mathbf{1}_{1\times n}\otimes\mathbf{b}_l^{feed})\nonumber\\
\!\! &\!\! \!\! &\!\! +\mathbf{1}_{1\times n}\otimes\mathbf{b}_l^{forward}
\end{eqnarray}
\noindent où $\mathbf{W}_l^{feed}\in\mathbb{R}^{E\times d_\mathbf{h}}$, $\mathbf{W}_l^{forward}\in\mathbb{R}^{d_\mathbf{h} \times E}$ et généralement $E > d_\mathbf{h}$.

En prenant en compte les couches de normalisation il est possible de déterminer le nombre de paramètres présents dans le réseau à l'aide de l'opération suivante\footnote{La dernière couche de \textit{pooling} n'a pas été prise en compte dans le calcul.}~:
\begin{eqnarray}
p = L(4d_\mathbf{h}^2+2d_\mathbf{h}E+9d_\mathbf{h}+E)+d_\mathbf{h}(|\mathcal{W}|+|\mathcal{E}|+2)
\end{eqnarray}
\noindent où $|\mathcal{W}|$ est la taille du dictionnaire et $|\mathcal{E}|$ est le nombre maximum de tokens possible pour le \textit{positional encoder}. Par exemple, pour CamemBERT, $|\mathcal{W}|=32'005$ et $|\mathcal{E}|=514$. A l'aide de ces éléments il est facile de décrire la structure CamemBERT qu'on peut résumer dans le tableau~\ref{tab:camembert}.

\section{La distillation}
\label{sec:distillation}

Le domaine du \textit{Knowledge Distillation} (KD) est une discipline relativement récente introduite en 2006~\cite{CrRi2006} afin de contrer des modélisations de plus en plus complexes. C'est en 2015~\cite{GeOr2015} que l'approche est généralisée et l'intérêt est démontré sur des modèles de classification d'image (qui à l'époque était le domaine où l'on trouvait les plus gros réseaux de neurones, donc les plus coûteux). Le principe est très facile à formaliser : le but est d'entraîner un modèle moins coûteux en nombre de paramètres (dans le lexique du domaine KD il s'agit du modèle \textbf{étudiant}) en s'aidant d'un modèle de référence (modèle \textbf{professeur}).

C'est en 2019~\cite{SaDe2019} que la société HuggingFace introduit pour la première fois une technique de distillation sur un modèle de type Transformer, alias BERT. Il existe aujourd'hui d'autre approches qui ont été appliquées à ce type de modélisation (MobileBERT~\cite{SuYu2019}, MiniLM~\cite{WeFu2020} en sont deux exemples), mais DistilBERT est le premier à en montrer l'intérêt.

\subsection{Structure de l'étudiant}
Distil* a pour objectif de réduire le nombre de couches Transformers du modèle d'origine par un facteur 2. La structure est résumée dans la tableau~\ref{tab:camembert}. Pour l'initialisation de l’entraînement, les paramètres de la partie \textit{word embedding} et \textit{positional encoding} sont copiés ainsi que les paramètres d'une couche Transformer sur deux. Ceci permet de réduire la distance entre le minimum local objectif et celui du modèle de référence.
\begin{table}[htbp]
  \centering
  \begin{tabular}{|r|c|c|c|c|c|}
    \hline
    modèle & $L$ & $d_\mathbf{h}$ & $J$ & $E$ & $p$ \\ \hline
    CamemBERT & 12 & 768 & 12 & 3'072 & 110M \\ \hline
    DistilCamemBERT & 6 & 768 & 12 & 3'072 & 67,5M \\ \hline
  \end{tabular}
  \caption{Description du réseau CamemBERT et DistilCamemBERT.}
  \label{tab:camembert}
\end{table}

\subsection{Fonctions coût}

\begin{figure}[h!]
	\centering
	\includegraphics[scale=0.5]{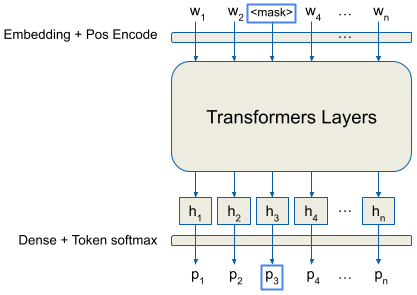}
	\caption{Schéma simplifié d'un réseau Transformer dans le cadre de DistilCamemBERT. Le cadrage bleu permet d'illustrer la correspondance entre un token masqué et la distribution de probabilité sur le dictionnaire correspondant pour une tâche de MLM.}
	\label{fig:structure_model}
\end{figure}

Dans cette partie, nous allons lister les différentes fonctions coût utiles à l'apprentissage de la modélisation. Afin d'aider à la compréhension et l'explication des différentes fonctions coûts qui composent la fonction coût d'apprentissage, nous avons réalisé la figure~\ref{fig:structure_model} qui illustre les éléments de la structure Transformer utilisée que ce soit pour le modèle professeur (CamemBERT) ou étudiant (DistilCamemBERT).\\
Nous posons $\mathcal{O}$ l'ensemble des observations d'apprentissage et $\mathcal{M}$ l'ensemble des tokens masqués. De même, l'exposant $s$ de $x^{s}$ fait référence à la partie étudiante et $t$ de $x^{t}$ au modèle professeur.

\textbf{Soft label :} cette fonction coût caractérise les méthodes de distillation sur les modèles de classification. Il s'agit d'exploiter la densité de probabilité sur l'ensemble du dictionnaire du modèle professeur et de rapprocher le comportement étudiant de celui-ci. La minimisation de la divergence de Kullback-Leibler sur chaque token permet de garantir un comportement semblable~:
\begin{equation}
SoftLabelLoss = \frac{1}{\sum_{k\in\mathcal{O}}n_k}\sum_{k\in\mathcal{O}}\sum_{0\leq i<n_k}D_{KL}(p_{k,i}^s\Vert p_{k,i}^t)
\end{equation}
\noindent Le principal reproche de cette fonction coût est de focaliser l'apprentissage du modèle élève sur la tâche d'estimation des tokens manquants (tâche MLM pour \textit{Masked Language Model}). Elle ne témoigne pas du fait que la modélisation est générique et que, par \textit{downstream}, le modèle s'adapte à des tâches très variées.

\textbf{Similarité Cosine :} a contrario, la dernière couche cachée représentant la dernière représentation latente des tokens est située en amont du classifieur. Une contrainte sur la colinéarité entre étudiant et professeur est donc appliquée :
\begin{equation}
CosineLoss = \frac{1}{\sum_{k\in\mathcal{O}}n_k}\sum_{k\in\mathcal{O}}\sum_{0\leq i<n_k}\frac{(\mathbf{h}_{k,i}^s)^T\mathbf{h}_{k,i}^t}{\Vert\mathbf{h}_{k,i}^s\Vert \Vert\mathbf{h}_{k,i}^t\Vert}
\end{equation}

\textbf{Tâche MLM :} il s'agit de la principale tâche d'entraînement du modèle professeur, elle est considéré comme étant générique. Il s'agit également d'une partie de l'apprentissage en autonomie pour le modèle étudiant, car ne s’appuyant pas sur les résultats du modèle professeur :
\begin{equation}
MLMLoss = \frac{-1}{\sum_{k\in\mathcal{O}}\vert\mathcal{M}_k\vert}\sum_{k\in\mathcal{O}}\sum_{i\in \mathcal{M}_k}p_{k,i}^w\log(p_{k,i}^s)
\end{equation}
\noindent où $p^w\in\lbrace 0;1\rbrace$ est le \textit{hard label} du mot masqué.

Finalement la fonction coût d'apprentissage est un mélange linéaire des trois précédentes. Le choix des coefficients a été choisi arbitrairement. L'idée était d'appliquer plus de poids à la tâche d'imitation des \textit{soft label} représentant, dans un contexte de classification, la fonction coût \textquote{classique} d'une distillation, puis de garder l'apprentissage MLM en tâche de fond. Pour finir, la fonction est calculée comme suit : 
\begin{eqnarray}
Loss\!\! &\!\! =\!\! &\!\! 0.5\times SoftLabelLoss + 0.3\times CosineLoss\nonumber\\ 
\!\! &\!\! \!\! &\!\! + \;  0.2 \times MLMLoss
\end{eqnarray}

\section{Résultats}
\label{sec:resultats}

Afin d'estimer la performance du modèle nous allons comparer la modélisation à CamemBERT dans 4 tâches types de ce genre de modélisation. En effet, le modèle est optimisé pour le moment pour de l'estimation de tokens cachés qui est une tâche générique. Il est possible via un \textit{fine tuning} d'adapter le modèle facilement à d'autres tâches plus utiles (\textit{downstream task}). Quand c'est possible nous utiliserons des modélisations CamemBERT déjà existantes présentes sur la plateforme de partage HuggingFace-Hub : Analyse de sentiment~\cite{Bl2020}, \textit{Natural Language Inference}~\cite{Do2021}, \textit{Question-Answering}~\cite{Et2020}.

\begin{figure*}[h!]
	\centering
	\includegraphics[scale=0.43]{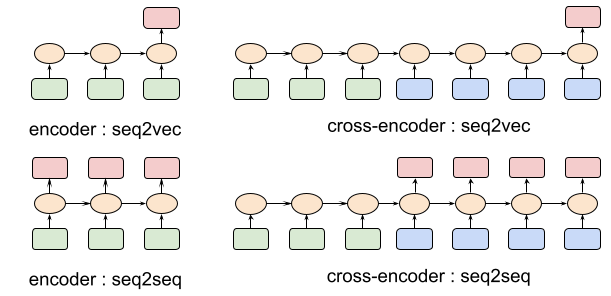}
	\caption{Différents types de structure classiques qu'on peut rencontrer en NLP. Les cases vertes (resp. bleues) représentent les tokens du premier texte (resp. second texte), les ovales représentent le processus de transformation d'attention et les cases en rouge la sortie du modèle qu'on souhaite exploiter.}
	\label{fig:structure}
\end{figure*}

Il est difficile d'être exhaustif dans l'évaluation des performances de ce type de modélisation, mais afin d'avoir une bonne vision globale des capacités de DistilCamemBERT nous avons choisit 4 tâches faisant intervenir 4 moyens différents d'utiliser l'information contenue dans le texte. Une vue des différentes approches est schématisée sur la figure~\ref{fig:structure}. 

L'ensemble des résultats sur les différentes tâches est disponible dans la tableau~\ref{tab:resultats} où la performance est mesurée avec la métrique f1-score.

\subsection{Analyse de sentiment}
\label{subsec:analysedesentiment}
Une tâche de classification de texte vient exploiter le premier token qui est dans le cas de BERT un token de classification, ou dans RoBERTa (CamemBERT) le token \lstinline{<p>}. Contrairement à ce que laisse croire le token, ce token n'est utilisé qu'en début d'entrée, il ne sépare ni les phrases, ni les paragraphes si plusieurs devaient être en entrée. Ainsi ce token est neutre et placé toujours au même endroit durant l’apprentissage, il contient donc l'information de l'ensemble du texte. Cette tâche va donc exploiter les informations \textit{intra} texte.

Il s'agit d'une classification bi-classe qui vise à estimer \textquote{positif} ou \textquote{négatif} des commentaires d'utilisateurs issus du site AlloCiné, donc très orientés critique de films, et d'Amazon, plus orientés produit et qualité de service.

Globalement on peut voir que les deux modélisations donnent des résultats très proches.

\subsection{Named Entity Recognition}
\label{subsec:ner}
Il s'agit d'une tâche de classification de tokens qui permet d'identifier des éléments d'un texte. A l’instar de la classification de texte, cette tâche exploite les informations \textit{intra} texte. L'application la plus représentative de cette tâche est la détection d'entités nommées (\textit{Named Entity Recognition}) ou simplement tâche NER.

Ici la modélisation reconnaît 4 classes : les personnes, les organisations, les lieux et les références culturelles nommées.

Pour mesurer les performances pour cette tâche, nous avons utilisé la définition du f1-score servant pour les applications de question-réponse, qui mesure la qualité de l'intersection des tokens classés et de ceux issus de la vérité terrain~\cite{RaZh2016}. Sur cette tâche également on peut voir que la performance de la modélisation DistilCamemBERT est proche de celle de son homologue CamemBERT.

\subsection{Natural Language Inference}
\label{subsec:ner}
Première tâche de \textit{cross-encoding} qui exploite les relations entre deux textes en entrée, elle est également connue sous le nom de \textit{Recognizing Textual Entailment} (RTE). Le premier texte est appelé prémisse et le second hypothèse. La tâche peut être décrite de la manière suivante :
\begin{equation}
P(pre=c\in\{contradiction,implication,neutre\}\vert hyp)
\end{equation}
\noindent avec respectivement $pre$ et $hyp$ pour prémisse et hypothèse. Cette tâche a la particularité de pouvoir créer des classifieurs \textquote{zero-shot}, c'est-à-dire des classifications sans entraînement :
\begin{equation}
P(hyp=i\in\mathcal{C}\vert pre)=\frac{e^{P(pre=implication\vert hyp=i)}}{\sum_{j\in\mathcal{C}}e^{P(pre=implication\vert hyp=i)}}
\end{equation}
\noindent avec $\mathcal{C}$ l'ensemble des classes possibles.

Il s'agit indéniablement d'une tâche plus complexe car elle exploite des informations \textit{extra} texte. On constate une perte de performance, bien que celle-ci ne soit pas dramatique.

\subsection{Question-Answering}
\label{subsec:qa}
La dernière tâche testée est celle de question-réponse correspondant à une structure de \textit{cross-encoding} de tokens labellisés. L'entrée se compose de deux textes, le premier est le \textquote{contexte} et le second la \textquote{question}. L'objectif est d'estimer le début et la fin de la réponse à la question dans le contexte via la labellisation des tokens. Au même titre que la tâche précédente, le modèle doit exploiter les relations \textit{extra} texte.

A l'instar de la tâche NER, la définition du f1-score utilisée est celle servant à mesurer les performances sur ce type de tâche~\cite{RaZh2016}. Cette tâche est très complexe et on constate une perte significative par rapport au modèle de référence.

\begin{table*}[htbp]
  \centering
  \begin{tabular}{|r|c|c|c|c|c|}
    \hline
    modèle & accélération & Sentiment (\%) & NER (\%) & NLI (\%) & QA (\%) \\ \hline
    CamemBERT & $\mathbf{\times}$1 & 95,74 & 88,93 & \textbf{81,68} & \textbf{79,57} \\ \hline
    DistilCamemBERT & $\mathbf{\times}$\textbf{2} & \textbf{97,57} & \textbf{89,12} & 77,48 & 62,65 \\ \hline
  \end{tabular}
  \caption{Comparaison des performances, mesurées en f1-score, entre le modèle CamemBERT et DistilCamemBERT sur différentes tâches communes à ce type de modélisation.}
  \label{tab:resultats}
\end{table*}

\section{Conclusions}
\label{sec:conclusion}

Avec une modélisation 2 fois plus petite il a été constaté que le temps d’inférence est bien 2 fois plus court.

Le modèle dans une approche \textit{encoder} fonctionne très bien et possède un très bon comportement, voire même un comportement un peu meilleur que la modélisation CamemBERT classique. Cette dernière remarque est toutefois à prendre avec des pincettes, la différence de performance n’étant pas significative et peut-être due à une différence de \textquote{qualité} du \textit{fine tuning} du modèle pré-entraîné.

Le modèle DistilCamemBERT montre ses limites dans les modélisations de type \textit{cross-encoding} avec une perte de 5 points notée sur la tâche de NLI et 15 points de perte sur la tâche de \textit{question-answering}. Les performances n’en demeurent pas moins bonnes. Mais là où, dans une modélisation de type \textit{encoder}, le gain est sans compromis (pas de perte de performance notable + gain en temps d’inférence), ici il s’agira d’un compromis entre temps d’exécution/performance que l’utilisateur final devra choisir en fonction de son objectif, de ses besoins et de ses contraintes.

Ces résultats ne sont pas surprenants. En effet, les modélisations à base d’un simple \textit{encoder} vont plutôt se concentrer sur la compréhension et la structure interne d’une phrase ou d'un texte, là où les \textit{cross-encoder} vont plus se concentrer sur la compréhension et la relation entre deux textes. On peut mettre en relation ce constat avec les analyses qui ont déjà était menées sur l’influence des couches~\cite{GaBo2019}. En effet, la compréhension des mots (morphosyntaxique et syntaxique) se fait plutôt dans les couches basses de la modélisation (relativement proches du \textit{word embedding}), alors que plus on monte dans des couches élevées, plus la modélisation acquiert une compréhension du texte (relation entre mots, entre phrases, etc.). Ainsi les structures avec un nombre de couches important sont plus disposées aux problématiques de \textit{cross-encoding}.

DistilCamemBERT présenté ici est \href{https://huggingface.co/cmarkea}{disponible} sur la plateforme HuggingFace-Hub en \textit{open-source}. Sous Python, il est très simple de charger le modèle, le code est représenté à la figure~\ref{fig:loadmodel}. Il en va de même pour toutes les \textit{downstream task} présentées, figure~\ref{fig:downstream}.

\begin{SaveVerbatim}{loadmodel}
from transformers import (AutoTokenizer,
                          AutoModel)
tokenizer = AutoTokenizer.from_pretrained(
    "cmarkea/distilcamembert-base"
)
model = AutoModel.from_pretrained(
    "cmarkea/distilcamembert-base"
)
model.eval()
\end{SaveVerbatim}

\begin{figure}[t]
  \centering
  \UseVerbatim[frame=single]{loadmodel}
  \vspace{-0.5cm}
  \caption{Chargement du modèle DistilCamemBERT.}
  \label{fig:loadmodel}
\end{figure}

\begin{SaveVerbatim}{downstream}
from transformers import pipeline
task = pipeline(task=..., model=*)
\end{SaveVerbatim}

\begin{figure}[t]
  \centering
  \UseVerbatim[frame=single]{downstream}
  \vspace{-0.5cm}
  \caption{Chargement d'une \textit{downstream task}, remplacer * par la tâche associée : \textquote{cmarkea/distilcamembert-base-\{sentiment, ner,~nli, qa\}} et le paramètre \textquote{task} par le nom de la tâche (se référer à la documentation de HuggingFace).}
  \label{fig:downstream}
\end{figure}

\end{document}